\documentclass[10pt,twocolumn,letterpaper]{article}

\usepackage{cvpr}      %

\def\paperTitle{Continuous-Time SO(3) Forecasting with\\ Savitzky--Golay Neural Controlled Differential Equations}

\def\authorBlock{
    Lennart Bastian\textsuperscript{1,2*} \qquad
    Mohammad Rashed\textsuperscript{1,2*} \qquad
    Nassir Navab\textsuperscript{1,2} \qquad
    Tolga Birdal\textsuperscript{3} \\
    \\
    \textsuperscript{1} Technical University of Munich \;
    \textsuperscript{2} Munich Center of Machine Learning \;
    \textsuperscript{3} Imperial College London
}

\makeatletter
\DeclareRobustCommand\onedot{\futurelet\@let@token\@onedot}
\def\@onedot{\ifx\@let@token.\else.\null\fi\xspace}

\makeatother

\newcommand{\SO}{\ensuremath{\mathrm{SO}(3)}}
\newcommand{\so}{\ensuremath{\mathfrak{so}(3)}}
\newcommand{\SE}{\ensuremath{\mathrm{SE}(3)}}

\newcommand{\R}{\mathbb{R}}

\newcommand{\bR}{\mathbf{R}}

\newcommand{\bJ}{\mathbf{J}}
\newcommand{\bomega}{\boldsymbol{\omega}}

\newcommand{\bA}{\mathbf{A}}
\newcommand{\bb}{\mathbf{b}}

\newcommand{\bx}{\mathbf{x}}
\newcommand{\bxtilde}{\mathbf{\tilde{x}}}

\newcommand{\hatop}[1]{\hat{#1}}

\newcommand{\phit}{\varphi(t)}

\newcommand{\ftheta}{f_\theta}

\newcommand{\zetatheta}{\zeta_\theta}
\newcommand{\expmap}{\mathrm{Exp}}
\newcommand{\logmap}{\mathrm{Log}}

\newcommand{\tk}{t_k}

\newcommand{\brho}{\boldsymbol{\rho}}
\newcommand{\brhok}{\boldsymbol{\rho}_k}
\newcommand{\bz}{\mathbf{z}}

\newcommand{\Xpath}{X}
\newcommand{\dX}{\mathrm{d}X}

\newcommand{\Nsamp}{N}

\newcommand{\bxtildek}{\mathbf{\tilde{x}}_k}

\usepackage{graphicx}	
\usepackage{amsmath}	
\usepackage{amssymb}	
\usepackage{booktabs}
\usepackage{times}
\usepackage{microtype}
\usepackage{epsfig}
\usepackage{caption}
\usepackage{colortbl}
\usepackage{enumitem}
\usepackage{tabularx}
\usepackage{xstring}
\usepackage{multirow}
\usepackage{xspace}
\usepackage{url}
\usepackage{subcaption}
\usepackage{footnote}                %
\usepackage{marvosym} %

\usepackage{cuted}

\usepackage{comment}
\usepackage{amsthm}
\usepackage{multirow}
\usepackage{refcount}

\usepackage[utf8]{inputenc}
\usepackage{makecell} 

\usepackage{xr-hyper}

\usepackage{thmtools}

\declaretheoremstyle[
  spaceabove=6pt,
  spacebelow=6pt,
  headfont=\normalfont\bfseries,
  notefont=\mdseries,
  notebraces={(}{)},
  bodyfont=\normalfont\itshape,
  postheadspace=1em
]{mytheoremstyle}

\declaretheorem[style=mytheoremstyle,name=Theorem]{theorem}

\declaretheorem[style=mytheoremstyle,name=Definition,numberlike=theorem]{definition}
\declaretheorem[style=mytheoremstyle,name=Lemma,numberlike=theorem]{lemma}
\declaretheorem[style=mytheoremstyle,name=Proposition,numbered=no]{proposition*}

\AtEndPreamble{
    \crefname{theorem}{Thm.}{Thms.}
    \crefname{proposition}{Prop.}{Props.}
    \crefname{definition}{Def.}{Defs.}
    \crefname{lemma}{Lem.}{Lems.}

    \creflabelformat{theorem}{(#2#1#3)}
    \creflabelformat{proposition}{(#2#1#3)}
    \creflabelformat{definition}{(#2#1#3)}
    \creflabelformat{lemma}{(#2#1#3)}
}

\renewcommand{\paragraph}[1]{{\vspace{0.6mm}\noindent \bf #1}.}

\definecolor{cvprblue}{rgb}{0.21,0.49,0.74}
\usepackage[pagebackref,breaklinks,colorlinks,allcolors=cvprblue]{hyperref}

\def\paperID{24} %
\def\confName{CVPRW 4DV}
\def\confYear{2025}

\begin{document}

\title{\paperTitle}
\author{\authorBlock}

\maketitle

{\let\thefootnote\relax\footnote{{
\vspace{-0.8cm}
\parbox{\linewidth}{%
 $^*$Equal Contribution. \quad
\raisebox{-0.2ex}{\Letter} \ : {\tt \scriptsize \{lennart.bastian,m.rashed\}@tum.de}
}}}}

\begin{abstract}
Tracking and forecasting the rotation of objects is fundamental in computer vision and robotics, yet $SO(3)$ extrapolation remains challenging as (1) sensor observations can be noisy and sparse, (2) motion patterns can be governed by complex dynamics, and (3) application settings can demand long-term forecasting. 
This work proposes modeling continuous-time rotational object dynamics on $SO(3)$ using Neural Controlled Differential Equations guided by Savitzky-Golay paths. 
Unlike existing methods that rely on simplified motion assumptions, our method learns a general latent dynamical system of the underlying object trajectory while
respecting the geometric structure of rotations.
Experimental results on real-world data demonstrate compelling forecasting capabilities compared to existing approaches.
\end{abstract}

\vspace{-5mm}
\section{Introduction}

Predicting the 3D rotation of objects is essential for numerous robotics applications, including object manipulation, grasping, and autonomous navigation. Tracking scenarios frequently involve temporary occlusions and detection failures, making accurate forecasting of rotational motion important for maintaining consistent object states.

A major challenge lies in representing the non-Euclidean structure of the rotation manifold $\mathrm{SO}(3)$, where even simple object motions can exhibit complex behavior~\cite{borisov2018rigid}. 
Unlike translational tracking in Euclidean space, rotational tracking must account for the non-linear nature of rotational motion governed by objects' moment of inertia. 
This becomes challenging when tracking systems produce noisy measurements due to occlusions or detection failures~\cite{fan2025benchmarks,park2008kinematic,gugushvili2012consistent}.

\begin{figure}[t]
    \centering
    \includegraphics[width=\columnwidth]{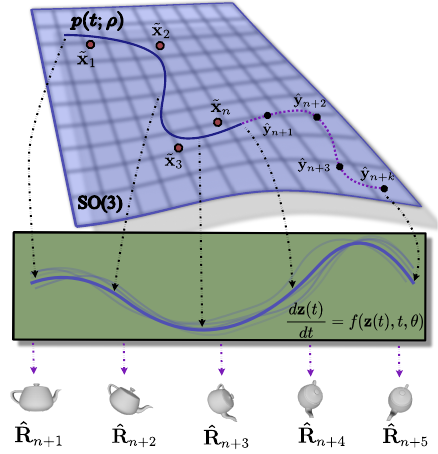}
    \caption{
      We propose Savitzy Golay neural Controlled Differential Equations (SG-nCDEs), learning a latent representation of \SO~dynamics with respect to a robust control path $\mathbf{p}(t; \cdot)$ with coefficients $\mathbf{\rho}$.
      The control path is constructed directly on the manifold $\mathrm{SO}(3)$ by filtering noisy input observations, enabling robust forecasting of future object orientations $\mathbf{\hat{y}}_k$.
      Our model derives the object dynamics directly from noisy pose estimates $\mathbf{\tilde{x}}_k \in \mathrm{SO}(3)$, without restrictive assumptions such as constant angular velocity.
    }
    \label{fig:cde_main_figure}
    \vspace{-5.0mm}
\end{figure}

We propose Savitzy Golay neural Controlled Differential Equations (SG-nCDEs) for rotational forecasting by modeling object dynamics through a latent dynamical system that respects the manifold structure of \SO. 
Unlike previous methods that rely on simplified assumptions like constant velocity or minimal torque \cite{mur2015orb}, our approach can predict rotational trajectories for complex object dynamics from noisy, sparse observations.
Our proposed method addresses existing limitations through a geometry-aware neural network that learns a continuous-time path in \SO, enabling long-term forecasting under noisy conditions. 
This supports advancements across multiple domains from robotic manipulation~\cite{marturi2019dynamic,wu2022grasparl}, autonomous navigation~\cite{ettinger2021large}, and tracking with sensor fusion~\cite{fan2025benchmarks,bosse2009continuous}.
Our contributions include:

\begin{itemize}
    \item A general data-driven approach for modeling \SO~object dynamics by integrating a latent dynamical system with neural controlled differential equations (nCDEs).
    \item Leveraging a robust integration control signal on the Lie group $\mathrm{SO}(3)$ that handles noisy input states while respecting the geometric structure of rotations.
    \item Experimental validation on real-world tracking data demonstrate superior performance, particularly in challenging scenarios with sensor noise and occlusions.
\end{itemize}

\section{Related Works}

We review relevant literature on $\mathrm{SO}(3)$ filtering for tracking, rotation representations, and trajectory forecasting methods.

\paragraph{$\mathrm{SO}(3)$ Filtering for Tracking} 
Robustly filtering rotational measurements is essential for tracking applications from video stabilization \cite{jia2014constrained} to robot localization \cite{tang2019white,wong2020data,talbot2024continuous} and sensor fusion~\cite{nirmal2016noise,busam2017camera}.
Filtering approaches using splines~\cite{kim1995general,lovegrove2013spline,haarbach2018survey} must preserve manifold properties of $\mathrm{SO}(3)$ to avoid numerical errors \cite{busam2017camera}, but their solutions cannot be computed in closed form~\cite{sommer2020efficient}.
Without explicit assumptions, splines diverge near interpolation boundaries \cite{persson2021practical,sommer2020efficient}, leading many tracking systems to extrapolate with constraints such as constant velocity or minimal torque \cite{mur2015orb}.

Savitzky-Golay filtering on $\mathrm{SO}(3)$ offers an efficient alternative by regressing polynomials in the tangent space and mapping back to the manifold via the exponential map \cite{jongeneel2022geometric}. 
This enables simultaneous angular velocity and acceleration estimation, recently exemplified for hand tracking~\cite{fan2025benchmarks}.
Other data-driven methods like Gaussian Processes~\cite{lang2014gaussian,lang2017computationally,lilge2024incorporating} offer principled uncertainty quantification but have limited representational capacity compared to deep neural networks and require careful kernel design for \SO.

\paragraph{Rotation Representations} 
Learning 3D rotations presents unique challenges. 
Classical approaches like Euler angles \cite{kundu20183d} or quaternions \cite{zhao2020quaternion} suffer from singularities or topological constraints that hinder direct regression~\cite{zhou2019continuity,bregier2021deepregression,geist2024learning} and backpropagation~\cite{chen2022projective,teed2021tangent}. 
This stems from $\mathrm{SO}(3)$ not being homeomorphic to any subset of 4D Euclidean space~\cite{zhou2019continuity}. 
Recent work demonstrates that these limitations significantly impact downstream tasks~\cite{zhou2019continuity,levinson2020analysis,geist2024learning}, motivating higher-dimensional representations like 6D~\cite{zhou2019continuity}, 9D~\cite{levinson2020analysis}, and 10D~\cite{peretroukhin2020smooth} parameterizations with better numerical stability.

\paragraph{Trajectory Forecasting} 
Neural Ordinary Differential Equations (neural ODEs) solve initial value problems using neural networks~\cite{chen2018neural}, with extensions like GRU-ODEs~\cite{rubanova2019latent,de2019gru} incorporating sequential data more effectively.
Neural Controlled Differential Equations (Neural CDEs)~\cite{kidger2020neural} enable more flexible conditioning by integrating the hidden state with a smooth signal.

Most forecasting approaches for tracking focus on predicting Euclidean position, with rotational forecasting receiving less attention despite its importance for complete 6D pose tracking.
Some existing works \cite{byravan2017se3} address transformation prediction directly from point cloud sequences, but do not address longer time horizons.
Others assume conservational motion which is inherently limiting \cite{mason2023learning}.

Our approach addresses these limitations by learning to forecast rotational trajectories directly from noisy pose data while respecting the geometric structure of $\SO$, and avoiding restrictive assumptions on the underlying trajectory.

\section{Preliminaries}

\paragraph{SO(3) Representation}
While positions and rotations of tracked objects each have three degrees of freedom, the space of rotations forms a non-commutative Lie group, making rotational tracking inherently more complex.
We begin by introducing key background and definitions.

\begin{definition}[Special Orthogonal Group]
\label{def:so3}
The special orthogonal group $\mathrm{SO}(3)$ represents the set of all rotations about the origin in $\mathbb{R}^3$:
\begin{equation}
\mathrm{SO}(3)=\left\{\mathbf{R} \in \mathbb{R}^{3 \times 3}: \mathbf{R}^{\top} \mathbf{R}=\mathbf{I}, \operatorname{det}(\mathbf{R})=1\right\}
\end{equation}
\end{definition}

\noindent The Lie algebra $\mathfrak{so}(3)$, the tangent space at the identity of $\mathrm{SO}(3)$, consists of real $3 \times 3$ skew-symmetric matrices.

\begin{definition}[Hat Operator]
\label{def:hat_operator}
Any $\bomega \in \mathbb{R}^3$ can be associated with an element $\hat{\bomega} \in \mathfrak{so}(3)$ via the hat operator:
\begin{align}
\hat{\bomega} = S(\bomega) = \begin{pmatrix}
0 & -\omega_3 & \omega_2 \\
\omega_3 & 0 & -\omega_1 \\
-\omega_2 & \omega_1 & 0 \\
\end{pmatrix}
\end{align}
\end{definition}

The time evolution of a rotation matrix $R(t)$ can be parameterized by $\dot{R} = R \hat{\bomega}$ for angular velocity $\bomega$ \cite{jongeneel2022geometric}, with $\mathfrak{so}(3)$ isomorphic to the tangent space $T_{\bR}\mathrm{SO}(3)$ at $\bR$ \cite{lee2017global}.

\section{SO(3) Savitzky-Golay Neural-CDEs}
\label{sec:method}

Our proposed rotational forecasting approach uses a neural CDE integrated with respect to a control signal on the manifold $\mathrm{SO}(3)$, constructed using a Savitzky-Golay filter that provides robust denoising while respecting the geometric structure of rotations. An overview can be seen in \cref{fig:cde_main_figure}.

\subsection{Neural Controlled Differential Equations}

\begin{definition}[Discrete Trajectories in $\mathrm{SO}(3)$]
\label{def:observations}
Let $\mathcal{T} = \{(t_k, \bxtilde_k)\}_{k=0}^{\Nsamp}$ be a sequence of observations of an object at discrete time points $t_k$, where $\bxtilde_k \in \mathrm{SO}(3)$ represents a noisy rotational measurement:
\begin{equation}
\nonumber
\mathcal{T} = \{(t_0, \bxtilde_0), (t_1, \bxtilde_1), \ldots, (t_{\Nsamp}, \bxtilde_{\Nsamp})\} \subset \R \times \mathrm{SO}(3)
\end{equation}
\end{definition}

Such a sequence can be obtained from a system tracking object pose. 
We aim to forecast the object's rotational trajectory during occlusions or for future time points $\hat{t} > t_N$.

\begin{definition}[Neural CDE \cite{kidger2020neural}]
\label{def:neural_cde}
Let $\Xpath: [t_0, t_n] \rightarrow \R^{v+1}$ be a control path, where $\Xpath_{t_i} = (t_i, \bx_i)$. 
A Neural CDE is defined as:
\begin{equation}
\label{eq:neural_cde}
\bz_t = \bz_{t_0} + \int_{t_0}^t \ftheta(\bz_s) \dX_s \quad \text{for } t \in (t_0, t_n]
\end{equation}
where $\bz_t \in \R^w$ is the hidden state, $\ftheta: \R^w \rightarrow \R^{w \times (v+1)}$ is a neural network, and $\bz_{t_0} = \zetatheta(t_0, \bx_0)$ is the initial condition encoded by a network $\zetatheta$.
\end{definition}

While existing approaches construct control paths using cubic splines \cite{kidger2020neural} or Hermite splines \cite{morrill2022choice}, these methods neither respect the geometric structure of $\mathrm{SO}(3)$ nor handle tracking noise effectively.

\subsection{Robust SO(3) Savitzky-Golay Regression}
\label{sec:robust_regression}

To better address noisy scenarios, we propose constructing an integration path using Savitzky-Golay filtering on $\mathrm{SO}(3)$ \cite{jongeneel2022geometric}, which regresses a polynomial in the Lie algebra and maps it back to the manifold via the exponential map.
Such a control path can then be used to integrate the neural CDE in \cref{eq:neural_cde} during training and inference.

\begin{definition}[Lie Algebra Polynomial]
\label{def:lie_algebra_polynomial}
We define a second-order polynomial in the Lie algebra $\mathfrak{so}(3)$ as:
\begin{equation}
p(t; \brho) := \brho_0 + \brho_1 t + \frac{1}{2}\brho_2 t^2 \in \mathfrak{so}(3)
\end{equation}
where $\brho = [\brho_0; \brho_1; \brho_2] \in \R^9$ are polynomial coefficients.
\end{definition}

\begin{definition}[Control Path on $\mathrm{SO}(3)$]
\label{def:control_path}
Given observations $\mathcal{T}$, we define our smooth control path $\phit \in \mathrm{SO}(3)$ as:
\begin{equation}
\phit = \expmap(p(t-\tk; \brhok))\bxtildek
\end{equation}
where $\tk$ is the anchor point time, $\bxtildek \in \mathrm{SO}(3)$ is the corresponding measurement, $\expmap: \mathfrak{so}(3) \to \mathrm{SO}(3)$ is the exponential map, and $\brhok$ represents polynomial coefficients at time $\tk$.
\end{definition}

\begin{definition}[Savitzky-Golay Filtering on $\mathrm{SO}(3)$]
\label{def:sg_optimization}
The optimal polynomial coefficients $\brhok$ at time $\tk$ are:
\begin{align}
\label{eq:sg_optimization}
\brhok = \arg\min_{\brho} \sum_{m=-n}^{n} \|&(\logmap(\bxtilde_{k+m}\bxtilde_k^{-1}) \nonumber \\
&- p(t_{k+m}-\tk; \brho))\|^2
\end{align}
where $2n+1$ is the window size, and $\logmap: \mathrm{SO}(3) \to \mathfrak{so}(3)$ maps rotation differences to the Lie algebra.

The solution reduces to the following linear system:
\begin{equation}
\brhok = (\bA^\top\bA)^{-1}\bA^\top\bb
\end{equation}
where $\bA \in \R^{3(2n+1) \times 9}$ and $\bb \in \R^{3(2n+1)}$ are constructed from time and rotation differences.
\end{definition}

To address boundary artifacts that can hinder forecasting \cite{schmid2022and}, we propose a (learned) weighted formulation:

\begin{lemma}[Weighted Savitzky-Golay Optimization]
\label{lemma:weighted_sg}
The weighted polynomial coefficients can be computed as:
\begin{equation}
\boldsymbol{\rho} = (\mathbf{A}^\top \mathbf{W} \mathbf{A})^{-1} \mathbf{A}^\top \mathbf{W} \mathbf{b}
\end{equation}
where $\mathbf{W}$ contains learnable parameters updated through gradient descent.
\end{lemma}

\subsection{Learning with SO(3) SG Neural-CDEs}

The Savitzky-Golay filter provides a smooth, bounded path on the SO(3) manifold that respects its geometric structure while effectively filtering measurement noise.
We can now integrate the differential equation in \cref{def:neural_cde} with the control path constructed in \cref{lemma:weighted_sg}.
The resulting SO(3) SG-nCDEs are a robust family of learned continuous-time extrapolation methods, approximating unknown object motion more generally than previous works. 

\paragraph{Learning Rotational Kinematics}
Consistent with the literature \cite{zhou2019continuity,bregier2021deepregression,geist2024learning}, we use the 9D rotational derivatives $\dot{R} = R \hat{\bomega}$ from \cref{sec:robust_regression} as integration paths for the latent state in \cref{eq:neural_cde}. The time-evolved hidden state is projected into the 6D representation \cite{zhou2019continuity}:
\begin{align}
    \mathbf{r_k}=\left(\mathbf{\nu}_1, \mathbf{\nu}_2\right) \in \mathbb{R}^{3 \times 2}
\end{align}
From this representation, we recover the $\SO$ prediction via Gram-Schmidt orthonormalization. During inference, we evolve the differential equation in \cref{eq:neural_cde} forward in time for all future points $k \in [t_{N+1}, t_{N+m}]$, and regularize with respect to the ground truth.

\begin{definition}[Learning Objective]
\label{def:learning_objective}
We optimize the latent parameters $\theta$ of the neural CDE by minimizing the Frobenius norm between predicted and ground-truth rotations:
\begin{equation}
\label{eq:general_loss}
\arg \min_\theta \sum_{k=1}^m ||\mathbf{\hat{y}}_k - \mathbf{x_k}||_F
\end{equation}
where $\mathbf{\hat{y}}_k \in \SO$ is the rotation matrix recovered from the neural CDE solution at time $t_k$ via GSO, and $\mathbf{x_k}$ represents the ground-truth observation.
\end{definition}

\begin{figure*}[t!]
    \includegraphics[width=\textwidth]{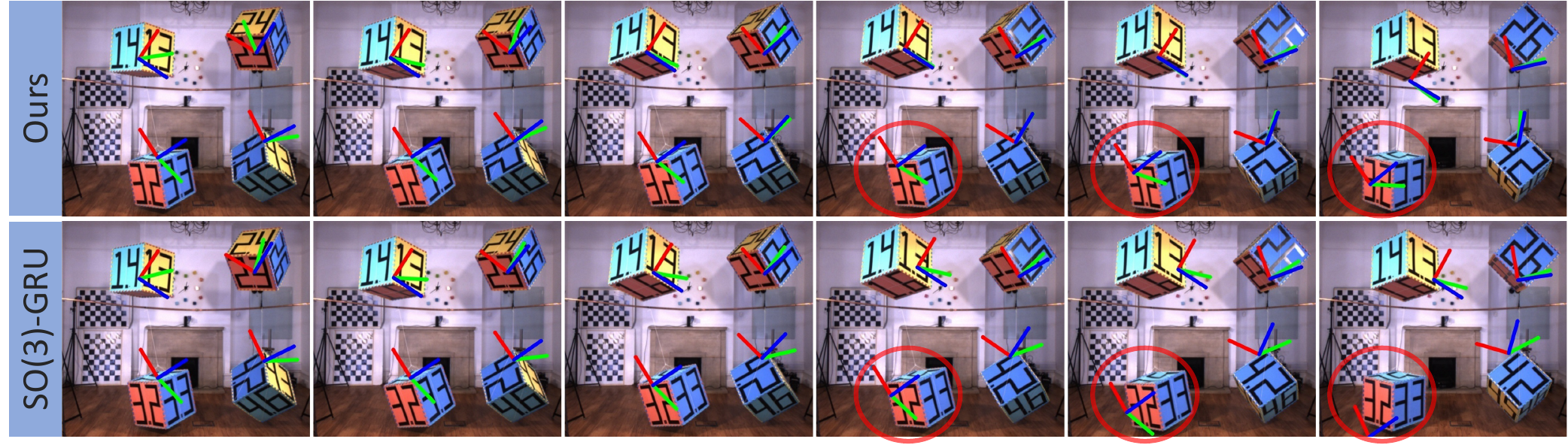}
    \vspace{-3mm}
    \caption{Qualitative comparison of the forecasting results produced by our method (top row) and \SO-GRU baseline (bottom row) on the OMD dataset. Methods are conditioned on the preceding timesteps (not visualized). Rotation is represented using 3D axis coordinates. 
    Ground truth observations are used for the translation component. Our method maintains more accurate alignment with the object rotation.}
    \label{fig:OMD_qualitative}
    \vspace{-3mm}
\end{figure*}

\section{Methodological Details}

Models are initially trained in simulation on diverse rotational patterns and then applied to real data to demonstrate their generalization capabilities to challenging tracking scenarios with occlusions and noise.
Simulation details are listed in the appendix.

\paragraph{Implementation Details}
We use Python version 3.12 and PyTorch \cite{paszke2019pytorch} for all implementations, along with \textit{torchdiffeq} \cite{torchdiffeq} for numerical integration using Dormand-Prince 4/5. Models are optimized using Adam~\cite{kingma2014adam}.

\paragraph{Evaluation}
During model evaluation, we measure the rotational geodesic error (RGE) \cite{huynh2009metrics} on a prediction trajectory with a varying length of up to $8$ samples:
\begin{equation}
\mathrm{RGE}(R_1, R_2) = 2 \arcsin\left( \frac{\| R_2 - R_1 \|_F}{2\sqrt{2}} \right)    
\end{equation}

\subsection{Emperical Evaluation}

We evaluate our method in various real-world tracking scenarios, including data fused from multiple irregularly sampled sensors. 

\paragraph{Oxford Motion Dataset}
We evaluate our approach on the Oxford Motion Dataset (OMD)~\cite{judd2019oxford}, which contains observations of four swinging boxes under different camera motions. Vicon tracking data is downsampled to 40Hz, and models are evaluated on predicting 0.3 future seconds.

As shown in Table~\ref{table:omd_dataset}, our SG-nCDE method consistently outperforms baseline approaches across different motion patterns. The qualitative results in~\cref{fig:OMD_qualitative} demonstrate that our method maintains better alignment with the actual box rotations than the \SO-GRU baseline, whose predicted rotations cause the projected coordinate system to drift significantly over time (to better visualize discrepancies, we represent object pose in the world coordinate system such that errors manifest as translations in pixel space).

\begin{table}[t]
\resizebox{0.99\columnwidth}{!}{
   \setlength{\tabcolsep}{7pt}
   \small
   \centering
  \begin{tabular}{l|ccc}
    \toprule
    Method & \makecell{Translation\\ Motion} & \makecell{Unconstrained\\ Motion} & \makecell{Static\\ Motion} \\
    \midrule
    \SO-nCDE & $6.49 \pm 9.42$ & $5.43 \pm 7.09$ & $7.82 \pm 12.19$ \\
    \SO-GRU & \textcolor{blue}{$3.04 \pm 1.29$} & \textcolor{blue}{$2.90 \pm 1.25$} & \textcolor{blue}{$2.83 \pm 1.26$} \\
    SG-nCDE (Ours) & \boldmath \textbf{$2.32 \pm 1.03$} & \boldmath \textbf{$2.30 \pm 1.02$} & \boldmath \textbf{$2.18 \pm 1.00$} \\
    \bottomrule
  \end{tabular}%
  }
  \caption{RGE (degrees) across different motion scenarios in the OMD dataset. \textbf{Best} and \textcolor{blue}{second-best} results are emphasized.}
  \label{table:omd_dataset}
  \vspace{-2mm}
\end{table}

\begin{figure}[t]
    \vspace{-1.0mm}
    \centering
    \includegraphics[width=\columnwidth]{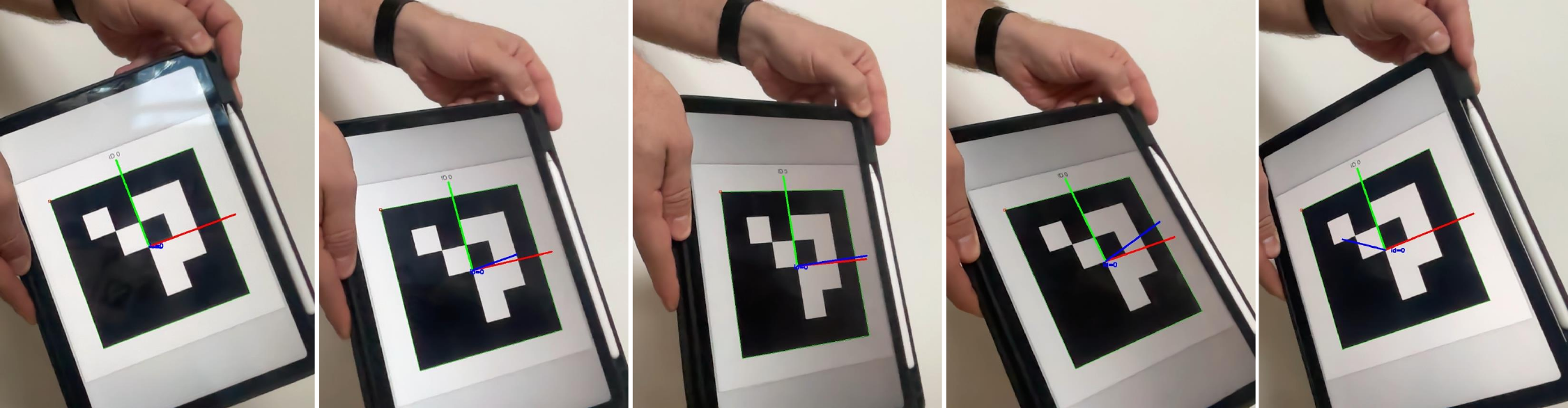}
    \caption{Evaluating the robustness of the proposed extrapolation methods for irregularly sampled extrapolation in the wild. The IMU signal from a tablet is synchronized with an external camera capturing the pose of a displayed ArUco marker.}
    \label{fig:sensor_fusion}
    \vspace{-1.0mm}
\end{figure}

\paragraph{Sensor Fusion}
We further evaluate our approach on data acquired with a tracking tablet. \SO measurements are acquired using an ArUco \cite{garrido2014automatic} marker with an external camera (30Hz) and onboard IMU (100Hz) (see~\cref{fig:sensor_fusion}).
This represents a challenging real-world scenario; two sensors are time-synchronized via Levenberg-Marquardt optimization and naively merged without additional filtering.
Quantitative results are depicted in~\cref{table:irregular_sampling}.
Our method outperforms baselines, demonstrating effectiveness on irregular, noisy signals from multiple sensors.

\begin{table}[t]
    \caption{Comparison of prediction errors across different models from irregularly sampled noisy sensor measurements. Results show mean ± standard deviation in rotational geodesic error (RGE).}
    \label{table:irregular_sampling}
    \centering
    \setlength{\tabcolsep}{4pt}
    \begin{tabular}{l|c}
    \toprule
    Model & Prediction Error \\
    \midrule
    \SO-nCDE & $75.48 \pm 9.39$ \\
    \SO-GRU & $11.39 \pm 0.73$ \\
    SG-nCDE (Ours) & $\mathbf{8.11 \pm 1.59}$ \\
    \bottomrule
    \end{tabular}
    \vspace{-2.0mm}
\end{table}

\vspace{-1mm}\section{Conclusion}
We propose a method for forecasting rotational motion in object tracking scenarios using Savitzky-Golay Neural Controlled Differential Equations. 
Our approach addresses challenging tracking problems where objects may be temporarily occluded or tracking data contains noise and inconsistencies.
By leveraging continuous-time latent dynamics and modeling the geometric structure on the manifold of rotations \SO, we achieve more accurate predictions than existing methods across two real-world scenarios.

\paragraph{Limitations and future work}
Our approach focuses on rotational forecasting in \SO, but tracking often requires predicting pose in \SE.
Extending our framework to the special Euclidean group \SE~to simultaneously predict position and orientation would be valuable for complete 6D object tracking.
However, this presents additional challenges such as filtering and modeling (or neglecting \cite{sommer2020efficient}) coupled rotational motion in the Lie algebra of twists \cite{haarbach2018survey}.

\clearpage

{
    \small
    \bibliographystyle{ieeenat_fullname}
    \bibliography{main}
}

\clearpage

\setcounter{page}{1}
\appendix

\maketitlesupplementary

\section{Experimental Details}

We leverage data from diverse simulated rotation trajectories to train our data-driven dynamics model.

\subsection{Rotational Kinematics Background}
\label{sec:training_in_simulation}

The motion of rotating rigid bodies follows well-established physical principles from classical mechanics. 
These principles enable us to simulate diverse rotational scenarios to train our method in a controlled environment.

\begin{definition}[Rotational Rigid Body Dynamics]
\label{def:rotational_dynamics}
The evolution of rigid body rotation is governed by:
\begin{align}
\dot{R} &= R \hat{\omega}, \\
\dot{\omega} &= \bJ^{-1} (\tau_{\text{ext}} - \omega \times \bJ \omega),
\end{align}
where $R(t) \in \SO$ is the orientation, $\omega(t) \in \mathbb{R}^3$ is the body-frame angular velocity, $\mathbf{J} \in \mathbb{R}^{3 \times 3}$ is the moment of inertia tensor characterizing the mass distribution, and $\tau_{\text{ext}} \in \mathbb{R}^3$ represents external torques.
\end{definition}

In real-world applications, strict conservation laws may not always hold due to dissipative forces and measurement noise. 
We therefore design our simulation framework to include both conservative (as in \cite{mason2023learning}) and non-conservative dynamics ($\tau_{\text{ext}} \neq 0$). This includes several key scenarios:

\begin{itemize}
    \item \textit{Freely Rotating:} Systems with no external torques where angular momentum is conserved, but gyroscopic effects can still produce complex non-linear motion
    \item \textit{Linear Control:} External torques following linear control laws, providing a testbed for simpler dynamics
    \item \textit{Configuration Dependent Torque:} External torques that depend on the orientation itself, simulating physical systems like magnetic dipoles in uniform fields
    \item \textit{Damped Motion:} Systems with dissipative forces that reduce angular momentum over time
\end{itemize}

To expose the models to diverse motion trajectories, we vary the moment of inertia tensor across simulations, creating rigid bodies with different principal axis orientations while maintaining similar underlying physics. 
Initial conditions are sampled according to \cite{shoemake1992uniform} with rejection sampling, avoiding degenerate cases with small initial angular velocity.

\section{Implementation Details}

We provide additional details regarding model implementation, specifically, the exact parameterization of the network in \cref{def:neural_cde}, and how we adapt the GRU and nCDE baselines to the manifold structure of \SO.
Hyperparameters for all models are tuned via a Bayesian parameter search, and optimal models are selected in our simulation environment (see \cref{sec:training_in_simulation}).

\paragraph{SG-nCDE}
To parameterize the function $f_\theta$ in \cref{def:neural_cde}, we use a neural network consisting of four linear layers and an Exponential Linear Unit (ELU) activation and a latent dimension of 100.
The initial encoding function $\zetatheta$ similarly consists of two linear layers and an ELU activation.
Following numerical integration of the function $f_\theta$, the latent representation is projected into the output dimension with two linear layers and an ELU.

\paragraph{\SO-nCDE}
We adapt the neural CDE baseline from \cite{kidger2020neural} to predict rotations in \SO.
Hermit cubic coefficients are directly constructed on the input 9D rotation representation \cite{zhou2019continuity} via backward differences.
Following \cite{kidger2020neural}, the method encodes an initial value $z_0$, which is then integrated forward in time using \textit{torchdiffeq} \cite{torchdiffeq} and Dormand-Prince 4/5 with respect to the constructed spline.
The latent representation is then decoded into the 6D rotation representation \cite{zhou2019continuity,geist2024learning}, and transformed to a rotation matrix with Gram-Schmidt orthonormalization (GSO).

\paragraph{\SO-GRU}
The GRU baseline consists of a multi-layer gated recurrent unit (GRU) RNN with three recurrent units, and a latent dimension of 250.
The GRU is similarly applied sequentially on each 9D rotation representation, yielding a 6D prediction converted to an element in \SO{} via GSO.

\subsection{On the Choice of Integration Path}

\begin{figure*}[t]
    \includegraphics[width=0.99\textwidth]{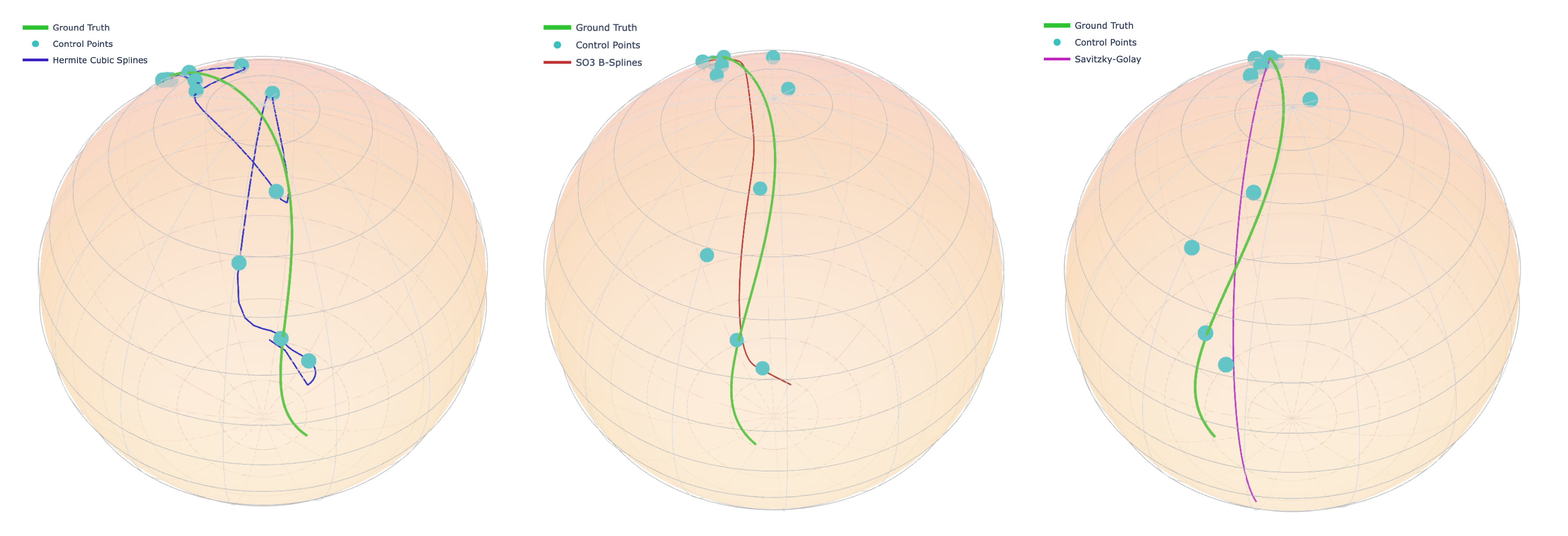}
    \vspace{-2mm}
    \caption{Comparison of interpolation methods on a particularly complex and noisy trajectory from the \textit{combined external force} scenario. From left to right: Hermite cubic splines with backward differences \cite{morrill2022choice}, \SO~B-splines \cite{sommer2020efficient}, and \SO~Savitzky Golay Filtering \cite{jongeneel2022geometric}.
    Trajectories represent interpolation and $t=0.2s$ extrapolation into the future.
    }
    \label{fig:spline_comparison}
\end{figure*}

Various control signals for neural CDEs have been proposed; most interpolation methods use cubic or higher order splines~\cite{kidger2020neural,morrill2022choice}. 
Kidger et al.~\cite{kidger2020neural} construct the integration path $\Xpath \in \mathcal{C}^2$ as a natural cubic spline over the input sequence.
While this guarantees smoothness of the first derivative, Hermite cubic splines with backward differences are preferred for CDE integration due to local control \cite{morrill2022choice}.
We observe that these choices do not obey the geometric structure of \SO; furthermore, they are not robust with respect to sensor noise interpolating the points erratically (see \cref{fig:spline_comparison}, left).
In general, such higher-order polynomials tend to diverge near the interpolation endpoints and must be extrapolated with some assumption, for instance, constant velocity \cite{persson2021practical}.
However, when dynamics are complicated, this assumption is overly simplifying, even more so than assuming conservation of energy as in \cite{mason2023learning}.

Efficient derivative calculations have been proposed for robust \SO~B-spline interpolation of trajectories in \SO~\cite{sommer2020efficient}.
These can be made suitable for extrapolation with, e.g., constant velocity extrapolation at the endpoint by constructing a nonlinear optimization to minimize residuals to the measurements, with additional smoothness near the endpoint.
We implement this in Ceres~\cite{Agarwal_Ceres_Solver_2022}, Basalt~\cite{usenko2018double}, and Sophus~\cite{sophus} to integrate Lie group constraints into the manifold optimization.
We use the efficient derivative computation of \cite{sommer2020efficient} and an additional first-order Taylor expansion using the computed Jacobians to include a constant velocity endpoint constraint defining the splines beyond their support interval.
While the splines nicely interpolate the noisy trajectory and can be used to extrapolate (see \cref{fig:spline_comparison}, center), they still exhibit bias towards endpoints.
Moreover, this approach is too computationally expensive to use during network training.
On a 24-Core Intel(R) Xeon(R) CPU, a single batch of 1000 trajectories takes $\approx 22$ seconds when maximally parallelized, a fraction of the trajectories we use for training.

On the other hand, the Savitzky-Golay filter provides a reasonable approximation at a low cost. 
As it merely requires solving a modest linear system, this means it can be differentiated through to learn a robust weighting suitable for extrapolation (see \cref{fig:spline_comparison}, right).

\section{Additional Background}

\subsection{SO(3) Savitzky-Golay Filtering \cite{jongeneel2022geometric}}
\label{sec:sg_filter_details}

The Savitzky-Golay filter on SO(3) solves a least-squares problem to estimate polynomial coefficients that best fit the noisy rotational data in the Lie algebra, as described in Definition 7.

The design matrix $\mathbf{A} \in \mathbb{R}^{3(2n + 1) \times 3(p+1) }$ takes the form of a Vandermonde matrix with time-shifted polynomial coefficients.
It is expanded with the Kroeneker product $\otimes$ and $\mathbf{I}_3$, the identity matrix in $\mathbb{R}^3$ such that $\mathbf{A} = \hat{A} \otimes \mathbf{I}_3$.
 $\hat{A} \in \mathbb{R}^{(2n + 1) \times (p + 1)}$ is defined as:

\[
\hat{A} = \begin{bmatrix}
1 & (t_{-n} - t_k)  & \cdots & \frac{1}{p} (t_{-n} - t_k)^p \\
1 & (t_{-n+1} - t_k) & \cdots & \frac{1}{p} (t_{-n+1} - t_k)^p \\
\vdots & \vdots & \vdots & \vdots \\
1 & (t_n - t_k) & \cdots & \frac{1}{p} (t_n - t_k)^p
\end{bmatrix}
\]

$\bb$ is constructed by rotational differences in the lie algebra and has the form:
\[
\bb = \begin{bmatrix}
\logmap(\bxtilde_{k-n}\bxtilde_k^{-1})^\vee \\
\vdots \\
\logmap(\bxtilde_{k+n}\bxtilde_k^{-1})^\vee
\end{bmatrix}
\]

$(\cdot)^\vee: \so \to \R^3$ is the inverse of the hat operator that maps skew-symmetric matrices to vectors.

This formulation brings several advantages over conventional interpolation with splines. 
It filters robustly in the region of a point (with an arbitrary support window) and can be solved without an iterative optimization.
Moreover, the derivatives are smooth up to the polynomial order, allowing us to integrate the latent dynamical system of a neural CDE.

\subsection{Exponential and Logarithmic Maps}

\begin{definition}[Exponential Map on SO(3)]
\label{def:exp_map}
The exponential map $\expmap: \so \to \SO$ is a surjective mapping from the Lie algebra $\so$ to the Lie group $\SO$ defined as:
\begin{equation}
\expmap(\boldsymbol{\xi}) = \sum_{n=0}^{\infty} \frac{1}{n!}\boldsymbol{\xi}^n = \mathbf{I} + \boldsymbol{\xi} + \frac{1}{2!}\boldsymbol{\xi}^2 + \frac{1}{3!}\boldsymbol{\xi}^3 + \cdots
\end{equation}
where $\boldsymbol{\xi} \in \so$ is a skew-symmetric matrix. 

Geometrically, if $\boldsymbol{\xi} = \hatop{\mathbf{v}}$ for some $\mathbf{v} \in \R^3$ with $\|\mathbf{v}\| = \theta$, then $\expmap(\boldsymbol{\xi})$ represents a rotation by angle $\theta$ about the axis $\mathbf{v}/\|\mathbf{v}\|$.
\end{definition}

\begin{definition}[Logarithmic Map on SO(3)]
\label{def:log_map}
The logarithmic map $\logmap: \SO \to \so$ is the local inverse of the exponential map, retrieving the corresponding Lie algebra element from a rotation matrix:
\begin{equation}
\logmap(\mathbf{R}) = \boldsymbol{\xi} \in \so
\end{equation}
such that $\expmap(\boldsymbol{\xi}) = \mathbf{R}$. This mapping is well-defined for all $\mathbf{R} \in \SO$ where the rotation angle $\theta$ satisfies $0 \leq \theta < \pi$.
\end{definition}

\end{document}